\documentclass[letterpaper, 10 pt, conference]{ieeeconf}  

\IEEEoverridecommandlockouts                              
\usepackage{graphicx}
\graphicspath{ {./images/} }
\usepackage[table,xcdraw]{xcolor}
\usepackage[english]{babel}
\usepackage[caption=false]{subfig}
\usepackage{color}
\usepackage{amsmath, amscd, amssymb}
\usepackage[utf8]{inputenc} 
\usepackage[T1]{fontenc}    
\usepackage{hyperref}       
\usepackage{url}            
\usepackage{booktabs}       
\usepackage{amsfonts}       
\usepackage{nicefrac}       
\usepackage{microtype}      
\usepackage{lipsum}
\usepackage{bbm}
\usepackage{amsmath, bm}
\usepackage{multirow}
\usepackage{epsfig}
\usepackage{amsmath}
\usepackage{mathtools}
\usepackage[]{algorithm2e}
\usepackage{algorithmic}
\usepackage{amssymb}
\usepackage{verbatim}
\usepackage{lipsum}
\usepackage{graphicx}
\usepackage{cuted}
\usepackage{cite}
\usepackage{ulem}
\usepackage{url}
\usepackage{soul}

\usepackage{amsmath}
\DeclarePairedDelimiter{\norm}{\lVert}{\rVert}

 \usepackage{tikz}
 \usetikzlibrary{shapes.geometric, arrows}
 \tikzstyle{startstop} = [rectangle, rounded corners, minimum width=1cm, minimum height=1cm,text centered, text width=2cm,draw=black, fill=red!30]
\tikzstyle{arrow} = [thick,->,>=stealth]
\tikzstyle{process} = [rectangle, minimum width=1cm, minimum height=1cm, text centered,text width=2cm ,draw=black, fill=orange!30]
\tikzstyle{code} = [rectangle, minimum width=1cm, minimum height=1cm, text centered,text width=2cm, draw=black, fill=green!30]

\title{\LARGE \bf
PT-MMD: A Novel Statistical Framework for the Evaluation of Generative Systems}

\author{Alexander Potapov$^{1}$, Ian Colbert$^{1}$, Ken Kreutz-Delgado$^{1}$, Alexander Cloninger$^{2}$, and Srinjoy Das$^{3**}$
\thanks{*This work was not supported by any organization}
\thanks{$^{1}$A. Potapov, I. Colbert and K. Kreutz-Delgado are with the Department of Electrical and Computer Engineering,
        University of California San Diego}%
\thanks{$^{2}$A. Cloninger is with the Department of Mathematics and Halicio{\u g}lu Data Science Institute, University of California San Diego}%
\thanks{$^{3}$S. Das is with the Department of Mathematics, University of California San Diego}%
\thanks{$^{**}$Corresponding author; email: s2das@ucsd.edu}%
}

\begin{document}

\maketitle
\thispagestyle{empty}
\pagestyle{empty}

\begin{abstract}

Stochastic-sampling-based Generative Neural Networks, such as Restricted Boltzmann Machines and Generative Adversarial Networks, are now used for applications such as denoising, image occlusion removal, pattern completion, and motion synthesis. In scenarios which involve performing such inference tasks with these models, it is critical to determine metrics that allow for model selection and/or maintenance of requisite generative performance under pre-specified implementation constraints. In this paper, we propose a new metric for evaluating generative model performance based on $p$-values derived from the combined use of Maximum Mean Discrepancy (MMD) and permutation-based (PT-based) resampling, which we refer to as PT-MMD. We demonstrate the effectiveness of this metric for two cases: (1) Selection of bitwidth and activation function complexity to achieve minimum power-at-performance for Restricted Boltzmann Machines; (2) Quantitative comparison of images generated by two types of Generative Adversarial Networks (PGAN and WGAN) to facilitate model selection in order to maximize the fidelity of generated images. For these applications, our results are shown using Euclidean and Haar-based kernels for the PT-MMD two sample hypothesis test. This demonstrates the critical role of distance functions in comparing generated images against their corresponding ground truth counterparts as what would be perceived by human users.

\end{abstract}

\section{INTRODUCTION}

Generative Models refer to models of probability distributions that can be used to encode the probability distribution of high dimensional data. Evaluating and comparing the performance of such models is an area of significant interest~\cite{pros_cons}. In this context, Maximum Mean Discrepancy (MMD) has been previously proposed as a training and evaluation metric for Generative Adversarial Networks (GANs) \cite{gan_eval,dziugaite2015training}. 
In this paper we propose a methodology for evaluation of inference tasks on pre-trained Generative Models using the framework of two sample hypothesis testing. 
The test statistic is formed by using kernel-based Maximum Mean Discrepancy where the distance function in the kernel is constructed using both conventional Euclidean and perceptual metrics. Our hypothesis testing framework uses permutation-based (PT-based) resampling to generate $p$-values as a measure of comparison between training data from a ground truth distribution (T) and that obtained by performing inference on a generative model of the data (G). Our proposed approach does not rely on raw MMD scores and, provided that the null hypothesis of distribution equality is true, has equal capability to distinguish between T and G versus that between T and T'. Here T' indicates a new set of training data obtained independently of $T$ and whose statistics are the same as the original data T.
The PT-MMD framework can be utilized for tasks such as comparing quality within a family of models, e.g. GANs, and facilitating selection of hyperparameters, such as bitwidths and sigmoidal activation function complexity, for performing inference on hardware realizations of generative neural networks, e.g. RBMs.\footnote{Bitwidth is specifically known to be linearly correlated with power consumption in RBMs~\cite{chihyin}.} 

\section{PROCEDURES}

A novel aspect of our proposed framework is addressing the limitations of conventional MMD kernels by comparing the performance of PT-MMD with both Euclidean and Haar-transform based kernels for evaluating image quality. Zhao et. al show that Euclidean distance based metrics used in many generative machine learning frameworks have perceptual limitations ~\cite{zhao2017loss}.
In juxtaposition, the Haar-transform based distance metric has been shown to be  strongly correlated with human opinion scores of image quality when compared to Euclidean distance~\cite{haarpsi,haar1910}. We construct our non-Euclidean MMD kernel by using {\it only} the directional components of the Haar-transform and demonstrate the advantage of this kernel versus Euclidean based kernels in our PT-MMD hypothesis test framework. Our hypothesis testing framework is built on permutation-based resampling of data. This approach is advantageous as it does not require {\it a priori} knowledge of the null distribution, which for finite samples is dataset specific, hence the desire to estimate it with an empirical approximation~\cite{kernelTST}. The following sections detail the mathematical background used in constructing PT-MMD, followed by our methodology and results for inference tasks using RBMs and GANs.

\section{Construction of PT-MMD}

We consider the setting where i.i.d. data samples  $X$ and $Y$ are observed from probability distributions $p_\text{\rm data}$ and $p_\text{\rm gen}$ respectively where the two datasets $X$ and $Y$ are considered independent. For the purpose of quantifying generative model performance our interest lies in the following general framework for two-sample testing:


\begin{align*}
H_0: p_\text{\rm data} = p_\text{\rm gen} \\
H_1: p_\text{\rm data} \neq p_\text{\rm gen}
\end{align*}


For the most general case the observed data samples $X$ and $Y$ have no point-to-point correspondence and arise from high-dimensional distributions $p_\text{\rm data}$ and $p_\text{\rm gen}$. In addition, in most cases of practical interest it may not be possible to make distributional assumptions on the data. Given these considerations our hypothesis testing framework is constructed using a permutation-based version of the Maximum Mean Discrepancy test (MMD) which is a kernel-based two-sample test that has been proposed in~\cite{kernelTST}. The following subsections describe some mathematical details of PT-MMD and the associated distance metrics that are used for the MMD kernel.

\subsection{MMD: Maximum Mean Discrepancy}

Maximum Mean Discrepancy is a measure of discrepancy between two distributions generally defined as:

\vspace{-0.4cm}
\begin{equation}
\label{eq.MMD_basic}
\text{MMD}(p,q;F) =  \underset{f \in F} {sup} \int \ f(v)\left(p(v)-q(v)\right) \ dv\
\end{equation}

\noindent where $F$ denotes a certain family of integrable functions and $p, q$ denote the two distributions under comparison. Kernel-based formulations of MMD as distance measures for high-dimensional distributions $p$ and $q$ have been proposed in \cite{kernelTST} where the function class F consists of all functions s.t. $|| f ||_H \leq 1$, where $||\cdot||_H$ indicates the norm of the Hilbert space associated with the reproducing kernel $k$ where $k(x_1, x_2) : X$ x $X \mapsto R$ is positive definite. Such Reproducing Kernel Hilbert Spaces will be referred to henceforth as RKHS. Specifically, suppose the positive definite kernel is $k(x, y)$, then using Eq~\ref{eq.MMD_basic}, the (squared) RKHS population MMD can be written as:

{\footnotesize \begin{equation}
\begin{aligned}
\label{eq.MMD_pop}
\text{MMD}^2(p,q) = \int\int k(v,w) (p(v) - q(v))(p(w)-q(w)) dv dw
\end{aligned}
\end{equation}
}

\noindent The MMD defined by Eq~\ref{eq.MMD_pop} is a metric when $H$ is a universal RKHS which satisfies the requirement that the kernel $k$ is continuous. For such universal RKHS a key property is \cite{kernelTST}:

{\small
\begin{equation}
\label{eq.MMD_univ}
\text{MMD}[F, p, q] = 0 \iff p=q
\end{equation}
}

\noindent The finite sample estimate of the population MMD stated above is defined as below where $\bm{x} \in X \sim p_{\rm data}$ and $\bm{y} \in Y \sim p_{\rm gen}$:

{\small
\begin{equation*}
\begin{aligned}
\text{MMD}^2(X,Y) = \frac{1}{n^2} \sum_{x,x^{'} \in X} k(x,x^{'}) \ +  \\ \frac{1}{m^2} \sum_{y,y^{'} \in Y} k(y,y^{'}) \\ -  \frac{2}{nm} \sum_{x \in X, y \in Y} k(x,y) 
\end{aligned}
\end{equation*}
}

\noindent We use this finite sample estimator of MMD along with the universality property defined in Eq~\ref{eq.MMD_univ} in our formulation of PT-MMD as a discrepancy measure between two high-dimensional distributions $p_{\rm data}$ and $p_{\rm gen}$.\\

\subsection{Distance Metrics}

The construction of MMD as a discrepancy measure between the two distributions $p_\text{\rm data}$ and $p_\text{\rm gen}$ is critically dependent on the distance function $D$ which is used in the kernel function $k$. For a Gaussian kernel this can be expressed between two data samples (typically high dimensional) $x \in X \sim p_\text{\rm data}$ and $y \in Y \sim p_\text{\rm gen}$ as below:

\begin{equation}
\label{eq.MMD_kernel}
k(x,y) = e^{-\frac{D(x,y)}{2\sigma^2}}
\end{equation}

\noindent In this paper we use two distance metrics for the MMD Gaussian kernel which are described as below:\\

\subsubsection{Euclidean Distance}

The standard construction of MMD uses an Euclidean ($L_2$) distance metric for the kernel where the distance is as defined below. Here Z is the dimensionality of an image data sample $x \in X$ (same for $y \in Y$):
$$
 D_E(x, y) = \norm{x-y}^2 = \sum_{i=1}^Z (x_i-y_i)^2
$$

For image datasets this can be performed by first constructing a 1-dimensional vector of pixel values and then computing $D_E$ as above.\\

\subsubsection{Haar Distance}

Our second construction is based on using components of the Haar transform \cite{haar1910} in order to create a distance metric that allows for perceptual similarity measurement between image datasets. When transforming an image with Haar wavelets, four outputs are derived which include the details on the horizontal, vertical, and diagonal components of the image along with an average component based on summing pixel values in a given neighborhood \cite{song2006wavelet}.\\ 

For our purposes we construct the Haar distance in two steps: 
\begin{itemize}
    \item Using the convolutional Haar transform on the original image $x$  to construct the vector $h_1(x)$ which is the concatenation of the horizontal, vertical and diagonal components of the transform as described above\\
    \item Using the average of $h_1(x)$ as the input image $x_1$ for the Haar transform and concatenate the horizontal, vertical and diagonal components of this second transform to construct the vector $h_2(x_1)$. We construct the vector $h(x)$ by stacking $h_1(x)$ and $h_2(x_1)$ into a single vector for an input image $x$ as defined below, where $Z$ is the dimensionality of an image data samples $x \in X$ and $y \in Y$.
\end{itemize}

\begin{equation}
\begin{aligned}
h_{\rm full}(x) &= [\ h_1(x) \ h_2(x_1)\ ]\\
h_{\rm full}(y) &= [\ h_1(y) \ h_2(y_1)\ ]\\
D_H(x, y) &= \norm{h_{\rm full}(x) - h_{\rm full}(y)}^2 \\ 
&= \sum_{i=1}^Z (\ h_{\rm full}(x_i) - h_{\rm full}(y_i) \ )^2
\end{aligned}
\end{equation}

The computation data flow for the Haar distance is shown in Figure 1. An example of how the Haar components are derived from a sample image drawn from the MNIST dataset is shown in Figure \ref{fig:haar_transforms}. The Euclidean distance $D_E$ or Haar distance $D_H$ are used as the distance function $D$ in the MMD kernel as shown in Eq~\ref{eq.MMD_kernel}.

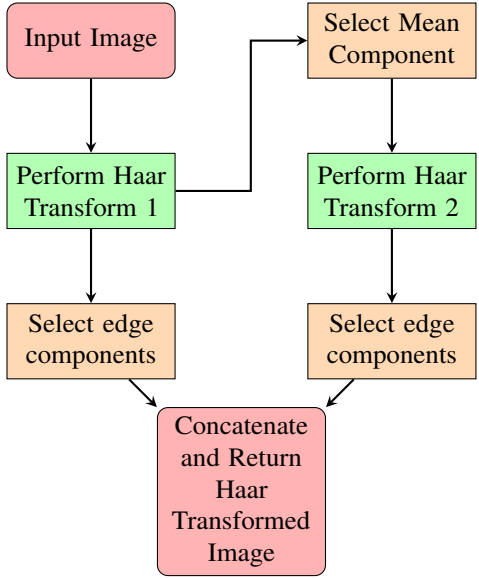
\begin{figure}[!ht]
\centering
\begin{tikzpicture}[node distance=2cm]
\node (start) [startstop] {Input Image};
\node (haar1) [code, below of=start] {Perform Haar Transform 1};
\node (mean1) [process, right of=start, xshift=2cm] {Select Mean Component};
\node (haar2) [code, below of=mean1] {Perform Haar Transform 2};
\node (edges) [process, below of=haar1] {Select edge components};
\node (edges2) [process, below of=haar2]{Select edge components};
\node (stop) [startstop, below of=edges, xshift=2cm] {Concatenate and Return Haar Transformed Image};

\draw [arrow] (start) -- (haar1);
\draw [arrow] (mean1) -- (haar2);
\draw [arrow] (haar2) -- (edges2);
\draw [arrow] (haar1) -- (edges);
\draw [arrow] (haar1.east) -- +(1,0) |- node[pos=0.25, anchor=east] {} (mean1);

\draw [arrow] (edges) -- (stop);
\draw [arrow] (edges2) -- (stop);

\end{tikzpicture}
\label{fig:flowchart}
\caption[Flowchart of Haar Pre-processing]{Flowchart of Haar Pre-processing}

\end{figure}

\begin{figure}[ht]
    \centering
  \subfloat[\label{fig:no_haar}Original Digit]{%
       \includegraphics[width=0.27\linewidth]{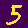}}
    \hfill
  \subfloat[\label{fig:one_haar}One Transform]{%
       \includegraphics[width=0.27\linewidth]{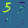}}
    \hfill
  \subfloat[\label{fig:two_haar}Two Transforms]{%
        \includegraphics[width=0.27\linewidth]{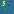}}
  \caption{Perceptual Image Representation Via Haar Transforms. In the transformed images, top left is the averaged variant, top right is the vertical detail, bottom left is the horizontal detail, and bottom right is the diagonal detail. The Haar distance metric is constructed using only the horizontal, vertical and diagonal components. \footnotesize{\textit{(Adobe Acrobat may be required to view this figure properly.)}}}%
  \label{fig:haar_transforms}%
\end{figure}

\subsection{PT: Permutation Testing}

In order to establish a statistically meaningful metric for comparing images obtained from two datasets we use the framework of two sample hypothesis testing. The test statistic in this case is the MMD score between the distributions $p_\text{\rm data}$ and $p_\text{\rm gen}$ based on their samples $\bm{x} \in X \sim p_\text{\rm data}$ and $\bm{y} \in Y \sim p_\text{\rm gen}$, respectively. In this context, the following considerations assume key importance. First, the null distribution of our MMD-based test statistic based on Euclidean or Haar distance kernels is not known {\it a priori}. Moreover, in the general case this null distribution will be dataset specific. Therefore, the most effective way to generate $p$-values in such a hypothesis testing scenario is by resampling the available data. In our work, we use permutation-based resampling to construct the full PT-MMD framework.

\begin{algorithm}
 \KwData{$\bm{x}$, drawn from $p_{\text{\rm data}}$, and $\bm{y}$, drawn from $p_\text{\rm gen}$}
 \KwResult{$p$-value of the similarity of $p_\text{\rm data}$ and $p_\text{\rm gen}$}
 \begin{algorithmic}
 \REQUIRE $\theta = MMD(\bm{x, y})$\;
 \REQUIRE count = 0\;
 \REQUIRE N = no. of permutations\;
 \While{have iterated less than N times}{
  z = pool($\bm{x, y}$)\;
  randomly sample z, get two sample sets $\bm \hat{x}$ and $\bm \hat{y}$\;
  $\hat{\theta} = MMD(\bm{\hat x, \hat y})$\;
  \If{$\hat{\theta} > \theta$}{
   count = count + 1
   }
 }
 \RETURN$\text{pval} =\frac{\text{count}}{N}$
 \end{algorithmic}
 \caption{Permutation Testing}
 \label{alg:pt}
\end{algorithm}

The key steps in the permutation algorithm with MMD as the test statistic is shown in Algorithm \ref{alg:pt}. Over a sufficient (N=250) number of permutations, we first compute the MMD score $(\hat \theta)$ of the two original sets of image data samples $\bm{x} \sim p_\text{\rm data}$ and $\bm{y} \sim p_\text{\rm gen}$. Following this, we pool the samples $\bm x$ and $\bm y$ into a single set $\bm z$. Then we derive a new set of two samples $\bm{\hat x}$ and $\bm{\hat y}$ by random sampling from this set $\bm z$. The MMD score $\hat \theta$ is then calculated for this permutation. Repeating this process $N$ times, we count how many of the permutations got a higher MMD score than that over the original sample sets $(\bm{x, y})$ and use this result to obtain a $p$-value for the test. In this manner the obtained $p$-value provides a measure of closeness of the two distributions $p_\text{\rm data} , p_\text{\rm gen}$.

\section{PT-MMD as used on MNIST-RBM}

Restricted Boltzmann Machines (RBMs) ~\cite{RBM} are a class of generative stochastic artificial neural networks that can learn a probability distribution over a given set of inputs. The RBM is a bipartite graph that consists of two layers - one visible layer vector $\bm v$, where the states of the units in this layer are driven by the input data, and one hidden layer vector $\bm h$. When trained as a generative model, the RBM can learn to capture the geometry of high dimensional data based on its energy function derived from the Boltzmann distribution~\cite{haykin2009neural}. For such networks it can be shown that a necessary and sufficient condition for sampling from the Boltzmann distribution is to sample each neuron with a sigmoidal probability law which can be expressed as a function of the activities of all other connected neurons \cite{rojas2013neural}. Design of the sigmoidal function along with selecting the right bitwidth is therefore of critical importance when RBMs are used for generative inference tasks such as image denoising, interpolation and generation in real-time applications on low-power digital processors.\\

\subsection{Methods}

\begin{figure}[ht]
    \centering
  \subfloat[\label{fig:sigmoids_plots}Plots of True and Approximated Sigmoid Functions]{%
       \includegraphics[width=0.45\linewidth]{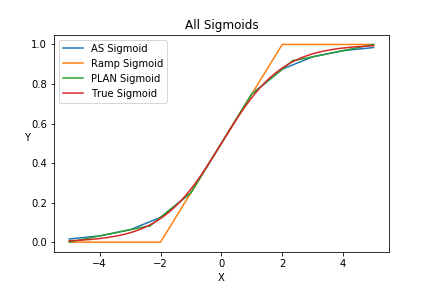}}
    \hfill
  \subfloat[\label{fig:sigmoids_error}Errors for Sigmoid Approximations versus True Sigmoid]{%
        \includegraphics[width=0.45\linewidth]{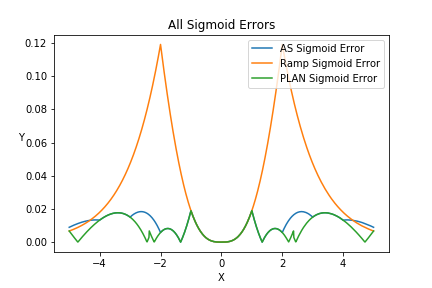}}
  \caption{This figure demonstrates the hardware sigmoids being considered in this experiment. We compare them to the true sigmoid. Note that as fidelity increases, so does hardware implementation cost~\cite{sigmoidpaper,chihyin}.}%
  \label{fig:sigmoids}%
\end{figure}

In this context, we demonstrate the use of PT-MMD in selecting bitwidth and sigmoidal function complexity in a RBM trained on the MNIST dataset~\cite{mnist}. As shown in Fig.~\ref{fig:sigmoids}, we consider three sigmoid approximations of varying complexity suitable for hardware implementations~\cite{sigmoidpaper}.

Given a user-specified significance level\footnote{Such significance levels can be based on pre-determined human opinion scores of image quality for given datasets.}  our design goal is to determine the optimal sigmoid and bitwidth, {\it i.e.} to achieve optimal power consumption, while at the same time meeting performance constraints which can be specified as significance levels for the test.

\subsection{Results}

\begin{figure}[t]
    \centering
  \subfloat[\label{fig:rbm_pval_euc}Euclidean PT-MMD]{%
       \includegraphics[width=0.45\linewidth]{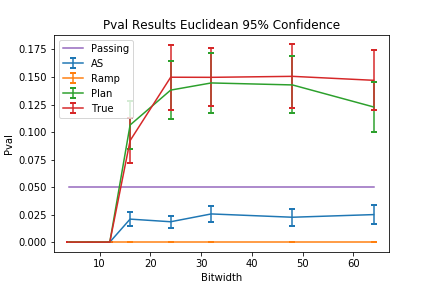}}
    \hfill
  \subfloat[\label{fig:rbm_pval_haar}Haar PT-MMD]{%
        \includegraphics[width=0.45\linewidth]{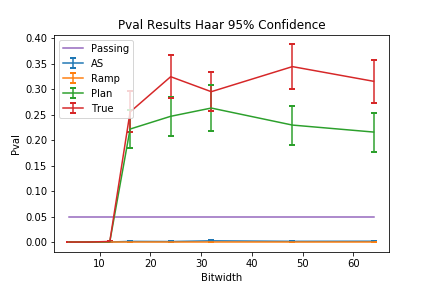}}
  \caption{$p$-value results for RBM Generated MNIST Data are shown above with the user specified significance level of 0.05 as a demonstrative example. In this application, the higher the $p$-value, the better the quality of the synthetic distribution. The trials are run at bitwidths of 4, 8, 12, 16, 24, 32, 48, and 64. A total of 100 Monte Carlo simulations were used for each bitwidth to capture the variation in its $p$-values. Results with an ideal sigmoidal function are shown as reference.}%
  \label{fig:rbm_pval}%
\end{figure}

Figure \ref{fig:rbm_pval_euc} shows the results of our PT-MMD test using the Euclidean-distance kernel for different bitwidths and sigmoid realizations. It can be seen that the PLAN sigmoid gives the highest $p$-values across bitwidths of 16 and higher which is consistent with the fact that PLAN  achieves the lowest error uniformly among all sigmoid approximations considered for our analysis as shown in Figure \ref{fig:sigmoids_error}. In the case of the Haar distance kernel, PLAN still gives the best results, however in this case near-zero $p$-values are seen for both Ramp and AS. Based on these results it can be concluded that the PLAN approximation with a 16-bit implementation is the lowest bitwidth implementation which can generate MNIST images of sufficient quality with high similarity to the original distribution.\footnote{It is also possible to extend our PT-MMD proposal with multiple hypothesis testing using an adjusted significance level (e.g. with Bonferroni correction) for applications where the generative performance will not vary monotonically with the design parameter as what is demonstrated here with a single bitwidth value. For example this can occur when the generative model is constructed using a mix of bitwidths and the design requirement is to select the optimal mix based on quality of the generated images.} In addition, given the fact that Haar distance is a perceptual metric, the PT-MMD test based on this kernel is able to better distinguish synthetic images generated using lower quality sigmoids like AS from the ground truth as seen in the contrast between $p$-values produced by AS with Euclidean versus Haar-based PT-MMD. 

\section{PT-MMD as used on LSUN-GAN}

\subsection{Methods}

GANs are an important class of Deep Generative Models that can be used to learn a mapping from a known iid distribution of inputs to a given data distribution. When used for generative inference such networks have several advantages over other methods including the ability to generate images of extremely high quality in a single step. Various type of GANs have been proposed in the literature using a variety of architectures and metrics (loss functions) used for training. Owing to the diversity of the design proposals and the potentially large number of applications of GANs for inference related to image processing tasks quantifying differences in generative performance based on rational metrics for critical considerations such as model selection is an area of significant interest.\\

\begin{figure}[ht]
    \centering
  \subfloat[\label{fig:ground_visuals}Ground Samples]{%
       \includegraphics[width=0.27\linewidth]{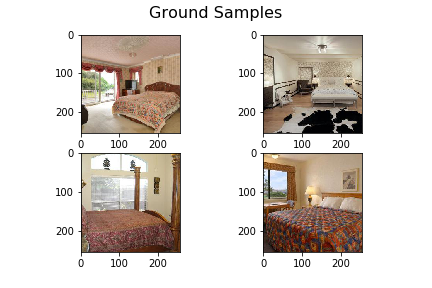}}
    \hfill
  \subfloat[\label{fig:wgan_visuals}WGAN Samples]{%
        \includegraphics[width=0.27\linewidth]{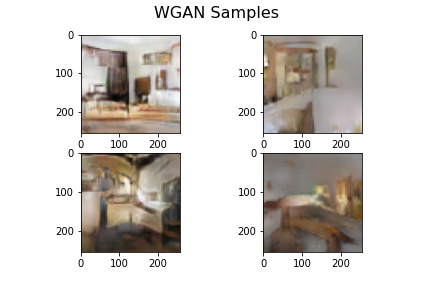}}
    \hfill
  \subfloat[\label{fig:pgan_visuals}PGAN Samples]{%
        \includegraphics[width=0.27\linewidth]{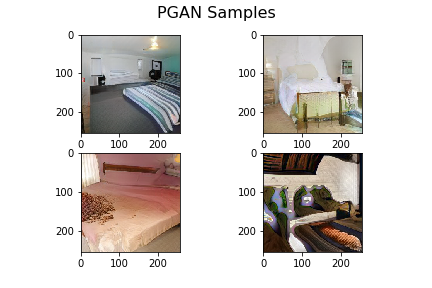}}
  \caption{We present example LSUN images generated by these GANs. Notably, we can see sharper edges in the PGAN as compared to the WGAN, acting as an early indicator of the advantages of PGAN to WGAN under the Haar transform. Being able to compare such results mathematically is critical to proper evaluation of these systems.}%
  \label{fig:gan_pictures}%
\end{figure}

We consider two previously proposed GAN models, namely PGAN and WGAN, trained on the LSUN dataset and apply our PT-MMD based tests to the generated images from these models~\cite{yu2015lsun,wgan,pgan,wgan_github}. 
The PGAN learns to create high quality images by starting with an extremely low resolution image generated from noise and progressively upsampling and improving that image. ~\cite{pgan}. 
We compare the PGAN versus the Wasserstein GAN (WGAN)~\cite{wgan} which was originally proposed as a GAN framework for learning a data distribution using the Wasserstein distance. Examples of LSUN data samples generated from these two GAN models are shown in Figure \ref{fig:gan_pictures}, subjectively demonstrating their generative quality differences. 
Our goal is to quantify this difference in generated image quality  using PT-MMD with both the Euclidean and Haar-based MMD kernels. 

\subsection{Results}

\begin{table}[ht]
\centering
\caption{GAN $p$-value Comparison (95\% Confidence)}
\begin{tabular}{||c c c||} 
 \hline
 Model & Euclidean & Haar \\ [0.5ex] 
 \hline\hline
 PGAN & $0.2316\pm0.0811$ & $0.00268\pm0.00126$ \\ 
 WGAN & $0.00\pm0.00$ & $0.00\pm0.00$ \\ 
 \hline
\end{tabular}
\label{fig:gan_pval}
\end{table}

The PT-MMD results are given in Table \ref{fig:gan_pval}. This table is based on 1000 samples of generated LSUN images compared against 1000 ground truth LSUN Images~\cite{yu2015lsun}. The WGAN fails the PT-MMD test at any significance level for both types of kernels. In contrast, the PGAN test fails to reject $H_0$ for the Euclidean distance kernel. In case of Haar, the observed $p$-values for PGAN are near, but non-zero, rejecting $H_0$, but still outperforming the WGAN.

\begin{figure}[ht]
    \centering
  \subfloat[\label{fig:pgan_euc_mmd}Euc PT-MMD PGAN]{%
       \includegraphics[width=0.45\linewidth]{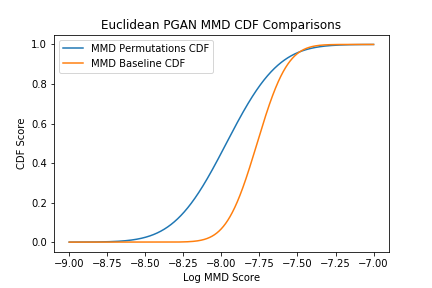}}
    \hfill
  \subfloat[\label{fig:pgan_haar_mmd}Haar PT-MMD PGAN]{%
        \includegraphics[width=0.45\linewidth]{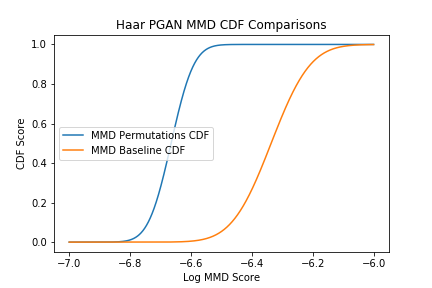}}
  \\
  \subfloat[\label{fig:wgan_euc_mmd}Euc PT-MMD WGAN]{%
       \includegraphics[width=0.45\linewidth]{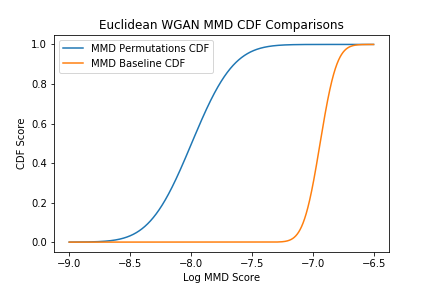}}
    \hfill
  \subfloat[\label{fig:wgan_haar_mmd}Haar PT-MMD WGAN]{%
        \includegraphics[width=0.45\linewidth]{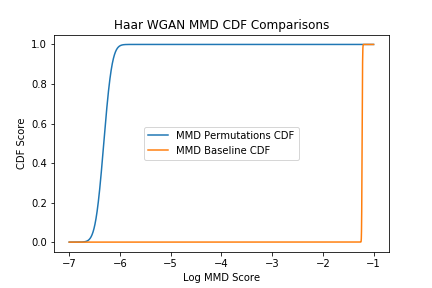}}  
  \caption{Cumulative Distribution Function (CDF) of obtained MMD scores in PT-MMD runs.
    For each graph, a decrease of the degree of separation in the two estimated CDFs corresponds to an increase in mean $p$-value of the PT-MMD test which signifies increasing closeness between ground truth and synthetic data distributions. Notably, under Haar PT-MMD, the separation between the two CDFs is much higher than under Euclidean PT-MMD, suggesting that the Haar distance is better at distinguishing synthetic distributions from true distributions.}%
  \label{fig:cdf_plots_gan}%
\end{figure}

The differences in PT-MMD performance are illustrated in more detail in Figure \ref{fig:cdf_plots_gan}, showing the separation between the CDF of the baseline MMD values of the generated data and the CDF of the MMD values obtained via permutations to estimate the null distribution. It can be seen that the extent of separation of the baseline and permutation CDFs is much lower for the PGAN resulting in higher $p$-values for the PT-MMD test as compared to the WGAN.

\section{CONCLUSIONS}

In this paper, we quantify the performance of Generative Models consistent with human perception. This is done by constructing a novel permutation variant of the 2-sample kernel MMD test based on Euclidean and Haar distances.
The applicability of PT-MMD is demonstrated for selection of design parameters such as bitwidth and sigmoid approximation for RBM hardware implementations, as well as automated model selection between GANs. For the datasets considered, we have also shown that the perceptual Haar-based PT-MMD is better at distinguishing images than PT-MMD parameterized by Euclidean distances.

\addtolength{\textheight}{-12cm}   





\section*{ACKNOWLEDGMENT}
This work was supported in part by NSF awards CNS-1730158, ACI-1540112, ACI-1541349, OAC-1826967, the University of California Office of the President, and the California Institute for Telecommunications and Information Technology's Qualcomm Institute (Calit2-QI). Thanks to CENIC for the 100Gpbs networks. AC was partially supported by NSF grant DMS-1819222. The authors would also like to thank Dhiman Sengupta at UCSD.


\bibliographystyle{plain}
\bibliography{references.bib}

\end{document}